\documentclass{article}

\usepackage[preprint]{neurips_2026}
\setcitestyle{numbers,square,comma}

\usepackage[utf8]{inputenc}
\usepackage[T1]{fontenc}
\usepackage{hyperref}
\usepackage{url}
\usepackage{booktabs}
\usepackage{amsfonts}
\usepackage{amsmath}
\usepackage{amssymb}
\usepackage{nicefrac}
\usepackage{microtype}
\usepackage[table]{xcolor}
\usepackage{graphicx}
\usepackage{enumitem}
\usepackage{multirow}
\usepackage{graphicx}
\usepackage{wrapfig}
\usepackage{caption}
\usepackage{algorithm}
\usepackage{algpseudocode}
\usepackage{placeins}
\usepackage{tcolorbox}
\tcbuselibrary{breakable}

\newtcolorbox{promptbox}[1]{
    breakable,
    colback=gray!5,
    colframe=black!45,
    title={#1},
    fonttitle=\bfseries,
    boxrule=0.5pt,
    arc=2pt,
    left=5pt,
    right=5pt,
    top=5pt,
    bottom=5pt
}

\usepackage{xspace}

\title{Learning New Tasks via Reusable Skills: Skill-Compositional Experts for Embodied Continual Learning}

\author{%
Shuaike Zhang$^{1}$ \quad Shaokun Wang$^{1}$ \quad Haoyu Tang$^{2}$ \quad Jianlong Wu$^{1,3}$ \quad Liqiang Nie$^{1,3}$\\
$^{1}$Harbin Institute of Technology, Shenzhen\\
$^{2}$Shandong University\\
$^{3}$Shenzhen Loop Area Institute\\
\texttt{zhangsk1218@gmail.com, wangshaokun@hit.edu.cn, tanghao258@sdu.edu.cn}\\
\texttt{wujianlong@hit.edu.cn, nieliqiang@gmail.com}
}

\begin{document}

\raggedbottom

\maketitle

\begin{abstract}
Embodied Continual Learning (ECL) aims to enable robots to continually acquire new manipulation tasks while retaining previously learned behaviors under closed-loop control. 
In ECL, feature drift can propagate through sequential decision-making under closed-loop control, turning representation changes into compounding behavioral deviations on previously learned tasks. 
A key challenge in ECL lies in structured skill reuse across continually evolving tasks, since existing methods primarily focus on skill learning without explicitly organizing them for coherent task execution. 
To address this issue, we propose SCE, a Skill-Compositional Experts framework for ECL. 
SCE builds a skill base via Compositional Skill Grounding (CSG), which decomposes task demonstrations into reusable skills. Based on this, Dual Execution-and-Transition Experts (DETE) enable new task learning through skill composition, where one branch ensures skill execution and the other supports transitions between skills for coherent behavior. 
Experiments on LIBERO benchmarks and real-world manipulation tasks show that SCE improves retention and overall task performance. Further feature drift analyses and ablation studies verify the effectiveness of our method. 
Project website: \url{https://eqcy.github.io/sce/}. 
\end{abstract}

\section{Introduction}

Continual learning (CL) is essential for building systems that can adapt to evolving tasks over time~\citep{parisi2019continual,delange2022continual}. Despite substantial progress, most existing CL methods are studied in settings that lack closed-loop interaction with dynamic environments, where data does not evolve through robot-environment interaction. 
In contrast, embodied robots learn through continuous interaction with the environment, where observations are coupled with actions and evolve over time, giving rise to Embodied Continual Learning (ECL)~\citep{lesort2020continual,jia2025hierarchical}. In this setting, robots must continually acquire new embodied tasks while preserving the ability to reliably execute previously learned behaviors under dynamically changing states.

\begin{figure}[t]
  \centering
  \includegraphics[width=\linewidth]{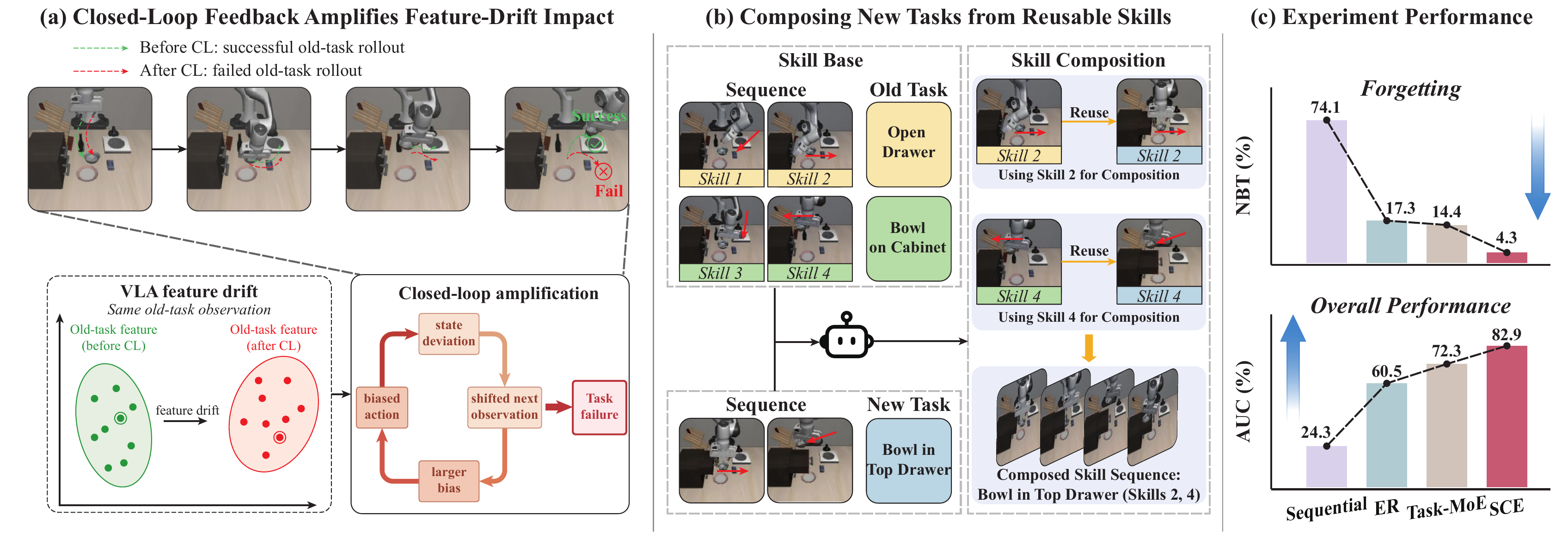}
  \caption{
      (a) Closed-loop control amplifies the impact of old-task feature drift.
      (b) New tasks can be composed from reusable skills in the skill base.
      (c) Forgetting and overall performance comparison on LIBERO-Goal.
  }
  \label{fig:introduction}
\end{figure}

In such interactive environments, catastrophic forgetting has direct behavioral consequences. As illustrated in Fig.~\ref{fig:introduction}(a), the feature representations of Vision-Language-Action (VLA) models can progressively drift toward new tasks. Under closed-loop control, this drift changes subsequent observations and states, so deviations can compound through sequential decision-making and degrade previously learned behaviors~\citep{ross2011reduction, spencer2021feedback}.
Therefore, ECL is crucial for improving the generalization and long-term adaptability of embodied robots, which are essential for real-world deployment~\citep{black2024pi_0,OpenVLA,RT-2}.

Existing ECL methods primarily mitigate catastrophic forgetting at the task or skill level. 
Task-level approaches preserve old knowledge through replay of previous task demonstrations~\citep{gao2021cril} or task-specific expert routing mechanisms~\citep{jia2025hierarchical}. 
However, they treat each task as an isolated learning unit, limiting the exploitation of shared patterns across tasks. This is suboptimal for embodied manipulation, where skill-level knowledge is essential for generalizing across diverse tasks.   
To address this limitation, skill-level methods~\citep{lee2024iscil} shift the focus to learning reusable skill primitives across tasks.  
Although effective, existing methods mainly focus on skill discovery~\citep{LOTUS} or evolution~\citep{wu2025stellarvla}, without explicitly modeling their organization for new task execution. 
As a result, this lack of structured skill organization leads to inconsistent skill reuse under sequential tasks, amplifying feature drift under closed-loop control. 
This highlights the need for structured skill reuse in continual learning.

Motivated by this insight, we propose a compositional view of ECL, where each task can be represented as a composition of reusable skills.
Based on this view, we introduce a Skill-Compositional Experts (SCE) framework for ECL.
As illustrated in Fig.~\ref{fig:introduction}(b), SCE maintains a skill base that stores reusable skills acquired from previous tasks, allowing new tasks to be learned through composition of relevant skills. 
Specifically, SCE consists of two components.
First, Compositional Skill Grounding (CSG) decomposes task demonstrations into reusable skills and organizes them into the skill base, enabling reuse of skill knowledge across tasks. 
Second, the Dual Execution-and-Transition Experts (DETE) module further augments the VLA action decoder with two complementary expert branches. 
The Execution Expert Branch models skill-specific action patterns for the execution of reusable skills, while the Transition Expert Branch models cross-skill transition patterns for skill switching during task execution. 
In addition, DETE employs the Adaptive Fusion Module to dynamically combine the outputs of the two branches for closed-loop action generation. 
Through this design, SCE maintains stable skill execution while flexibly handling skill transitions, thereby reducing feature drift during ECL. 
Comprehensive experiments on both simulated and real-world datasets demonstrate the advantages of SCE. 
In simulation, SCE achieves lower forgetting while maintaining stronger overall performance, as shown in Fig.~\ref{fig:introduction}(c).
Real-world robot experiments further verify its robustness in physical manipulation scenarios.
In-depth analyses further show that SCE effectively alleviates feature drift under closed-loop control. 
Our contributions are summarized as follows:

(1) We propose SCE, a Skill-Compositional Experts framework for ECL that enables continual task learning through composition of reusable skills, while preserving previously learned behaviors. 

(2) We introduce CSG and DETE. CSG decomposes task demonstrations into reusable skills and organizes them into a skill base, while DETE supports skill execution and cross-skill transitions for coherent skill composition. 

(3) Extensive experiments in both simulated and real-world manipulation environments demonstrate the effectiveness of SCE. 

\section{Related work}

\subsection{Continual learning}

Methods for addressing catastrophic forgetting in Continual Learning (CL) are commonly grouped into four categories:
Regularization-based~\citep{LwF,BiCL,Tpcil,Balance,OSAKA} approaches constrain parameter or feature updates to preserve old knowledge.
Architecture-based approaches~\citep{DER,mallya2018packnet,AANet,MGSVF} expand, prune, or reallocate model capacity to reduce interference among tasks.
Rehearsal-based approaches~\citep{rolnick2019experience,EndToEnd,LargeScale,CuriosityDriven,CausalEffect,AnchorReplay} store representative samples from previous tasks and replay them during current learning stages.
More recently, pre-trained model-based approaches~\citep{wang2022learning,wang2022dualprompt,Smith_2023_CVPR,InstancePrompt,PromptIncremental,EaSe,dia,peng2025cia} adapt strong backbones with prompts, adapters, or other parameter-efficient tuning strategies to improve the stability-plasticity trade-off.
Although these methods have achieved strong progress in mitigating catastrophic forgetting, they are mostly developed for settings where the evaluation inputs are fixed and independent of the model's predictions.
This assumption differs from Embodied Continual Learning (ECL), where a robot performs tasks under closed-loop control, such that each predicted action influences subsequent observations and states.
As a result, forgetting in ECL goes beyond performance degradation on static samples: errors can propagate over sequential action execution and compound into degraded embodied behaviors.
This makes ECL inherently more challenging than conventional CL and highlights the need for continual learning mechanisms tailored to closed-loop interaction.

\subsection{Embodied continual learning}

Embodied Continual Learning (ECL) extends CL to embodied robots that acquire sequential tasks through closed-loop interaction with the environment.
Recent VLA models and generalist policies have achieved strong progress in embodied manipulation~\citep{RT-2,OpenVLA,black2024pi_0,UniVLA,MoDE}. Nevertheless, these models remain susceptible to catastrophic forgetting when applied to ECL settings.
Existing ECL methods primarily mitigate this problem at either the task level or the skill level.
Task-level methods~\citep{liu2024tail,romer2026clare,topic} treat each task or task cluster as the basic unit for organizing continual knowledge.
For example, CRIL~\citep{gao2021cril} preserves prior task knowledge by reconstructing previous task trajectories with generative models.
MoILE~\citep{jia2025hierarchical} instead organizes knowledge through task clusters and routes inputs to hierarchical LoRA~\citep{hu2022lora} experts, reducing interference via task-aware expert selection.
M2Distill~\citep{M2Distill} preserves prior knowledge through multimodal distillation for lifelong imitation learning.
These methods demonstrate the effectiveness of task-level organization for reducing interference and preserving learned behaviors.
Nevertheless, task-level organization of continual knowledge may obscure shared behavioral structures across tasks.
Skill-level methods organize continual knowledge around finer-grained skill-oriented structures, enabling reusable behaviors to be shared across tasks~\citep{yao2025think,lee2025policy,zheng2025imanip,speci}.
For example, IsCiL~\citep{lee2024iscil} focuses on retrievable skills, where prototype-indexed skill adapters are retrieved for efficient continual task adaptation.
LOTUS~\citep{LOTUS} targets skill discovery, identifying temporal segments from demonstrations and organizing them into an expanding skill-policy library for manipulation.
Stellar VLA~\citep{wu2025stellarvla} promotes continually evolving skill knowledge, maintaining a task-skill knowledge space to guide VLA-based ECL.
T2S~\citep{zhang2025t2s} scales reusable knowledge through language-selected parameter tokens.
These methods show that organizing continual knowledge at the skill level can improve knowledge sharing and retention across tasks.
SCE builds on this skill-level view by explicitly modeling skill-specific execution and cross-skill transitions to compose reusable skills for new tasks.

\section{Method}
\label{sec:method}

\begin{figure}[t]
  \centering
  \includegraphics[width=\linewidth]{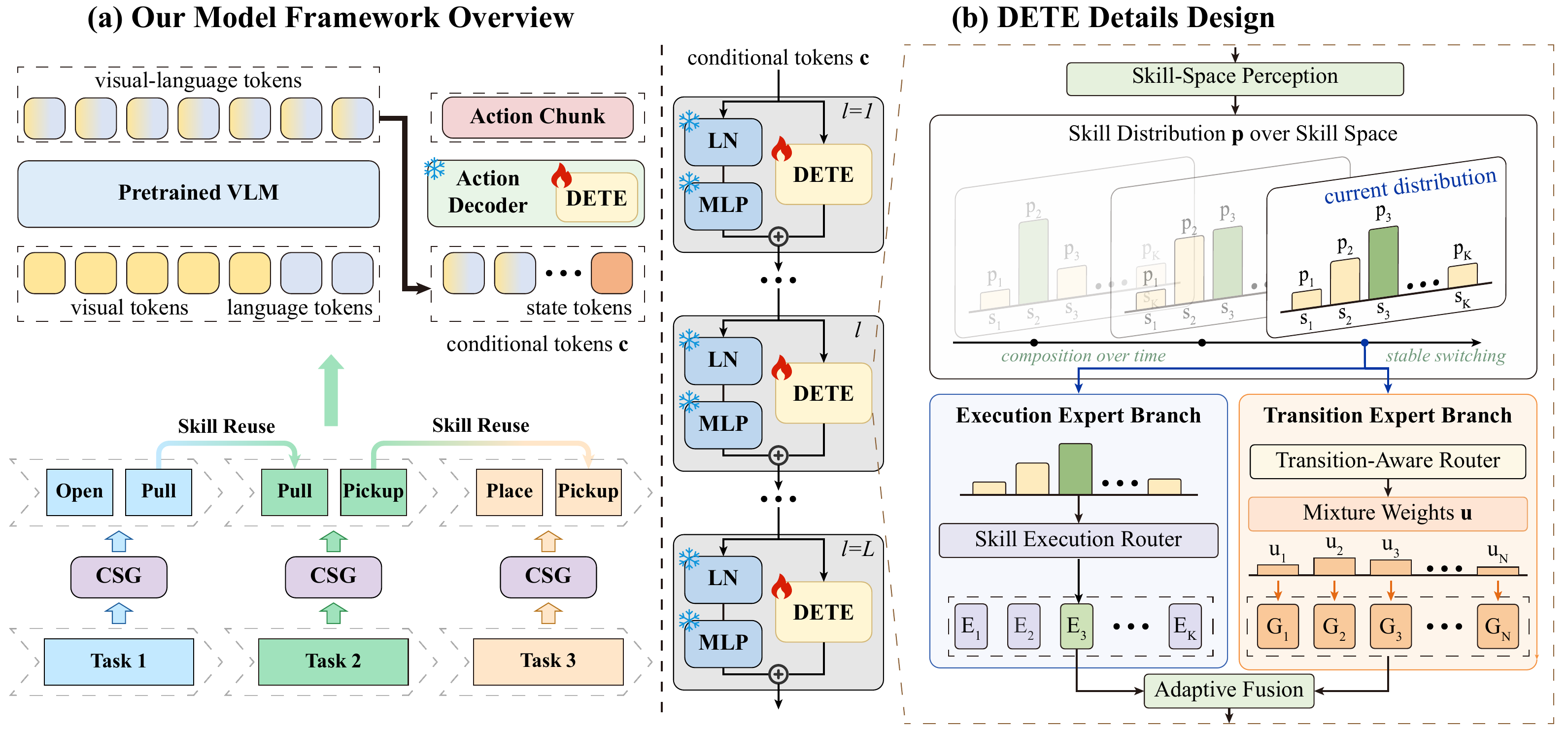}
    \caption{\textbf{Overall framework of the proposed method.}
    (a) The framework processes a sequential task stream by decomposing task demonstrations into reusable skills. DETE is integrated into the pretrained VLA action decoder to support skill-compositional embodied continual learning.
    (b) DETE combines skill-space perception with Execution and Transition Expert branches for skill-specific behavior modeling and coherent skill composition.}
  \label{fig:framework}
\end{figure}

As shown in Fig.~\ref{fig:framework}, SCE enables skill-compositional embodied continual learning by constructing a skill base and composing skills across tasks. It consists of CSG (Compositional Skill Grounding) and DETE (Dual Execution-and-Transition Experts). Specifically, CSG decomposes task demonstrations into reusable skills and organizes them into the skill base. DETE augments the VLA action decoder with two complementary expert branches that leverage reusable skills to model skill-specific behavioral patterns and cross-skill compositional knowledge, supporting execution and coherent skill composition across tasks.

\subsection{Problem definition}
\label{ssec:problem_definition}

In embodied continual learning, a VLA model is required to sequentially acquire new manipulation tasks while retaining previously learned ones. 
Let $\mathcal{D} = \{\mathcal{D}^{1}, \mathcal{D}^{2}, \dots, \mathcal{D}^{T}\}$ denote a stream of $T$ stages. 
At the $t$-th stage, the current dataset $\mathcal{D}^{t}$ and a bounded replay memory from previous stages are accessible. Each sample in $\mathcal{D}^{t}$ is represented as $x_{i} = \left( o_{i}, l_{i}, r_{i}, s_{i}, \mathbf{A}_{i} \right)$, where $o_{i}$ denotes the visual observation, $l_{i}$ is the language instruction, $r_{i}$ represents the robot state, $s_{i}$ denotes the associated skill annotation constructed as described in Sec.~\ref{ssec:skill_annotation}, and $\mathbf{A}_{i}=(\mathbf{a}_{i},\mathbf{a}_{i+1},\dots,\mathbf{a}_{i+H-1})$ denotes the corresponding $H$-step action chunk, with $\mathbf{a}_{i+h}$ denoting the robot action at future offset $h$ for $h=0,\dots,H-1$. 
Let $\mathcal{T}^{t}$ denote the set of tasks introduced at stage $t$, where $\forall i \neq j,\ \mathcal{T}^{i} \cap \mathcal{T}^{j} = \varnothing$. 
The model is sequentially trained on $\mathcal{D}^{1}, \mathcal{D}^{2}, \dots, \mathcal{D}^{T}$, and after training on $\mathcal{D}^{t}$, it is evaluated on all encountered tasks $\mathcal{T}^{1:t} = \bigcup_{k=1}^{t} \mathcal{T}^{k}$.

\subsection{Overall architecture}
\label{ssec:overall_architecture}

Fig.~\ref{fig:framework}(a) illustrates the overall architecture. 
Our method builds on a pretrained VLA model
and adds the DETE module to the action decoder. 
Given a sample $x=(o,l,r,s,\mathbf{A})$, the model predicts actions from $(o,l,r)$. The skill annotation $s$ provides auxiliary supervision for organizing the decoder around reusable skills.
During ECL at stage $t$, DETE is optimized only with the samples available at that stage; skills that appear only in future stages provide no positive skill supervision or action-generation data before their corresponding stages are reached.
We summarize the forward process as follows:
\begin{equation}
\label{eq:overall_forward}
\begin{aligned}
    \mathbf{z} &= \Phi(o,l), 
    & \mathbf{e} &= \Psi(r), \\
    \mathbf{c} &= [\mathbf{z}; \mathbf{e}],
    & \hat{\mathbf{A}} &= \Theta(\mathbf{c}),
\end{aligned}
\end{equation}
where $\Phi$ denotes the pretrained vision-language encoder, $\mathbf{z}$ is the visual-language token, $\Psi$ denotes the robot-state projection layer, $\mathbf{e}$ is the state token, and $\mathbf{c}$ is the conditional token formed by concatenating $\mathbf{z}$ and $\mathbf{e}$. 
The action decoder $\Theta$ predicts the corresponding $H$-step continuous action chunk
$\hat{\mathbf{A}}=(\hat{\mathbf{a}}_{0},\hat{\mathbf{a}}_{1},\dots,\hat{\mathbf{a}}_{H-1})$,
where $\hat{\mathbf{a}}_{j}$ denotes the predicted action at future offset $j$.

\subsection{Compositional Skill Grounding}
\label{ssec:skill_annotation}

This section details how CSG performs task decomposition over demonstrations to obtain reusable skills.
CSG proceeds in two phases: Phase 1 partitions each demonstration into temporally coherent segments according to robot-state dynamics, and Phase 2 grounds these segments with a Vision-Language Model (VLM) to induce reusable skills and update a skill base $\mathcal{B}$. The skill base is maintained across demonstrations and initialized as $\mathcal{B}=\emptyset$ before processing the demonstration stream.

\textbf{Phase 1: State-Aware Task Decomposition.}
Let $\mathcal{X}=(\mathcal{O},\mathcal{R},l)$ denote a raw task demonstration of temporal length $F$, where $\mathcal{O}=\{o_i\}_{i=1}^{F}$ is the observation sequence, $\mathcal{R}=\{r_i\}_{i=1}^{F}$ is the robot-state sequence, and $l$ is the language instruction.
Inspired by the keyframe intuition of Chain-of-Action~\citep{zhang2025chainofaction}, we define two state-based rules for temporal decomposition. Rule 1 marks a temporal boundary when the gripper state changes between adjacent timesteps. Rule 2 marks a temporal boundary when the end-effector motion trend changes abruptly.
CSG applies these rules to the robot-state sequence $\mathcal{R}$ and obtains a set of temporal boundary indices. These indices split the demonstration into $M$ contiguous temporal segments $\{\mathcal{X}^{(m)}\}_{m=1}^{M}$.
Each segment is defined as $\mathcal{X}^{(m)}=(\mathcal{O}^{(m)}, \mathcal{R}^{(m)},l)$, where $\mathcal{O}^{(m)}$ and $\mathcal{R}^{(m)}$ denote the corresponding observation and robot-state subsequences.

\textbf{Phase 2: VLM-Grounded Skill Induction.}
After temporal decomposition, CSG processes the temporal segments $\{\mathcal{X}^{(m)}\}_{m=1}^{M}$ sequentially using the current skill base $\mathcal{B}$.
For each temporal segment $\mathcal{X}^{(m)}$, the VLM~\citep{bai2025qwen3vl} grounds the segment observation $\mathcal{O}^{(m)}$ and the language instruction $l$ with reference to the current skill base $\mathcal{B}$, and proposes a candidate skill $\hat{s}^{(m)}$.
The candidate $\hat{s}^{(m)}$ is then used to retrieve matched skills from the skill base $\mathcal{B}$. If no existing skill in $\mathcal{B}$ matches $\hat{s}^{(m)}$, $\hat{s}^{(m)}$ is added to $\mathcal{B}$ as a new reusable skill. Otherwise, the matched skill is selected from $\mathcal{B}$.
The resulting skill assigned to $\mathcal{X}^{(m)}$ is denoted as $s^{(m)}$.
After all temporal segments are processed, CSG obtains the updated skill base $\mathcal{B}$ together with the segment-level skill assignments $\{(\mathcal{X}^{(m)}, s^{(m)})\}_{m=1}^{M}$.

The assignments $\{(\mathcal{X}^{(m)}, s^{(m)})\}_{m=1}^{M}$ are then converted into the sample-level skill annotations used in Sec.~\ref{ssec:problem_definition}.
Let $\mathcal{S}=\{s_i\}_{i=1}^{F}$ denote the resulting annotation sequence for demonstration $\mathcal{X}$. For any sample indexed by $i$ within temporal segment $\mathcal{X}^{(m)}$, we set $s_i=s^{(m)}$.
This yields a skill-annotated demonstration, while the skill base $\mathcal{B}$ is updated across the demonstration stream. CSG is applied to both simulated and real-world task demonstrations.

\subsection{Dual Execution-and-Transition Experts}
\label{ssec:adapters_design}

DETE adopts a dual-expert design with two complementary branches to support embodied continual learning through skill composition. The Execution Expert Branch focuses on modeling skill-specific behaviors, ensuring execution of individual skills. The Transition Expert Branch captures common knowledge across different skills, facilitating transitions among skills during composition. Together, these branches enable the skill-compositional process by combining skill-specific execution specialization with stable cross-skill transition modeling.

We attach DETE in parallel to the pretrained action decoder, while the decoder remains responsible for transforming encoded visual-language and robot-state features into executable robot actions. Since skill composition must ultimately be reflected in temporally coherent action sequences, this parallel interface allows DETE to provide skill-specific execution patterns and cross-skill transition structure to the decoder during action decoding. For one decoder layer, let $\mathbf{y}$ denote the feature entering DETE; we omit the layer index for clarity. Each expert is implemented as a bottleneck adapter with independent parameters.

\textbf{Skill-Space Perception.}
For skill composition along the action sequence, DETE needs a sample-level skill distribution that characterizes each input in terms of reusable skills and reflects changes in skill around transitions.
At stage $t$, this distribution is defined on the $K_t$-dimensional skill space $\mathcal{V}_t$ constructed from the current skill base $\mathcal{B}_t$, where $K_t=|\mathcal{B}_t|$. When CSG discovers a new skill, SCE appends one classifier output, its aligned Execution Expert, and the corresponding input column of the transition projection while preserving all existing skill indices and parameters. The Transition Expert pool remains fixed.

DETE derives the skill distribution $\mathbf{p} \in \mathbb{R}^{K_t}$ by mapping the conditional token $\mathbf{c}$ to the current skill space $\mathcal{V}_t$ and normalizing the result with a softmax function.
During training, this distribution is supervised by the automatically derived skill annotation:
\begin{equation}
\label{eq:skill_predictor_detail}
\mathbf{p} = \mathrm{softmax}(\mathrm{MLP}(\mathbf{c})), \qquad
\mathcal{L}_{\mathrm{skill}} = \mathrm{CE}(\mathbf{p}, \mathbf{s}).
\end{equation}
Here, $\mathbf{s}$ denotes the one-hot target encoded from the automatically derived skill annotation $s$, and $\mathrm{CE}(\cdot, \cdot)$ denotes the cross-entropy loss. At inference time, DETE obtains $\mathbf{p}$ from the observation, instruction, and robot state without requiring ground-truth skill annotations.

\textbf{Execution Expert Branch.}
During skill composition, each input is typically dominated by one skill that determines the local execution behavior. To stabilize skill-specific execution within a composed action sequence, DETE uses the skill distribution $\mathbf{p}$ to select the most relevant reusable skill.
We instantiate the Execution Expert pool $\mathcal{E}_t=\{E_k\}_{k=1}^{K_t}$, where $E_k$ is associated with the $k$-th dimension of the skill space $\mathcal{V}_t$. 

The Skill Execution Router routes the input to the skill expert with the skill distribution $\mathbf{p}$:
\begin{equation}
\label{eq:execution_branch}
  \hat{k}=\arg\max_k p_k,
  \qquad
  \Delta\mathbf{y}^{\mathrm{exe}}=E_{\hat{k}}(\mathbf{y}).
\end{equation}
Here, $p_k$ denotes the $k$-th component of $\mathbf{p}$, measuring the relevance of the input to the $k$-th skill dimension in $\mathcal{V}_t$; $\hat{k}$ denotes the selected skill index and $\Delta\mathbf{y}^{\mathrm{exe}}$ denotes the skill-specific output of the selected expert.
Under this design, execution-branch updates are confined to the expert associated with the currently involved skill, which reduces interference with unrelated skills and mitigates feature drift.
By assigning one Execution Expert to each skill, this branch provides dedicated capacity for stable skill-specific execution during composition.

\textbf{Transition Expert Branch.}
In real demonstrations, skill changes often extend over a temporal interval rather than occurring at a single instant. Around a skill transition, an input may contain cues from both the preceding skill and the subsequent skill. This ambiguity also appears at inference time, where ground-truth skill annotations are unavailable and the model must rely on the skill distribution $\mathbf{p}$. Therefore, the Transition Expert Branch treats $\mathbf{p}$ as a transition signal for modeling cross-skill dependencies during composition.
We instantiate an $N$-expert Transition Expert pool $\mathcal{G}=\{G_j\}_{j=1}^{N}$ to model cross-skill transitions.

The Transition-aware Router maps the conditional token $\mathbf{c}$ into a routing-feature space and injects the skill distribution $\mathbf{p}$ to obtain a transition-aware routing feature $\mathbf{h}$:
\begin{equation}
\label{eq:transition_routing_feature}
  \mathbf{h} = \mathrm{MLP}(\mathbf{c}) + W_{\mathrm{p}}\mathbf{p}.
\end{equation}
Here, $W_{\mathrm{p}}$ is a linear projection that maps $\mathbf{p}$ to the dimension of $\mathbf{h}$. Based on $\mathbf{h}$, this router computes mixture weights $\mathbf{u}\in\mathbb{R}^{N}$ over $\mathcal{G}$ and aggregates their outputs $\Delta\mathbf{y}^{\mathrm{tr}}$:
\begin{equation}
\label{eq:transition_branch}
  \mathbf{u}=\mathrm{softmax}(W_{\mathrm{u}}\mathbf{h}),
  \qquad
  \Delta\mathbf{y}^{\mathrm{tr}} = \sum_{j=1}^{N}u_j G_j(\mathbf{y}),
\end{equation}
where $W_{\mathrm{u}}$ is a linear projection and $u_j$ denotes the $j$-th component of the mixture weights $\mathbf{u}$. 
Since different inputs may require different degrees of skill-specific execution and cross-skill transition modeling, DETE employs an Adaptive Fusion Module to compute an adaptive balancing coefficient from the transition-aware routing feature $\mathbf{h}$:
\begin{equation}
\label{eq:dete_adapter}
    \alpha=\sigma(W_{\alpha}\mathbf{h}),
    \qquad
    \Delta\mathbf{y}
    =
    \alpha\Delta\mathbf{y}^{\mathrm{exe}}
    +
    (1-\alpha)\Delta\mathbf{y}^{\mathrm{tr}}.
\end{equation}
Here, $W_{\alpha}$ is a linear projection, $\sigma(\cdot)$ denotes the sigmoid function, $\alpha\in[0,1]$ is the adaptive balancing coefficient, and $\Delta\mathbf{y}$ denotes the combined DETE output.
The coefficient $\alpha$ adaptively balances skill-specific execution and cross-skill transitions through the transition-aware routing feature $\mathbf{h}$.

\subsection{Optimization objective}
The model is trained with the original action-generation objective and the auxiliary skill-distribution objective:
\begin{equation}
    \mathcal{L} = \mathcal{L}_{\mathrm{flow}} + \lambda \mathcal{L}_{\mathrm{skill}},
\end{equation}
where $\mathcal{L}_{\mathrm{flow}}$ is the flow-matching loss of the pretrained VLA policy~\citep{black2024pi_0}, $\mathcal{L}_{\mathrm{skill}}$ is the skill supervision loss defined in Eq.~\ref{eq:skill_predictor_detail}, and the hyperparameter $\lambda$ balances the two objectives. We use $\lambda=0.2$.

\section{Experiments}
\label{sec:experiment}

\begin{table*}[t]
    \centering
    \caption{
    Main results under ECL. Results are mean $\pm$ standard deviation over
    three independent training seeds. Prior results ($^\dagger$) are taken
    from Stellar VLA~\citep{wu2025stellarvla}.
    }
    \label{tab:main-results}
    \resizebox{\textwidth}{!}{
    \begin{tabular}{lcccccccc}
        \toprule
        \multirow{2}{*}{Method}
        & \multicolumn{4}{c}{LIBERO-Goal}
        & \multicolumn{4}{c}{LIBERO-Long} \\
        \cmidrule(lr){2-5} \cmidrule(lr){6-9}
        & FWT $\uparrow$ & NBT $\downarrow$ & AUC $\uparrow$ & Final SR $\uparrow$
        & FWT $\uparrow$ & NBT $\downarrow$ & AUC $\uparrow$ & Final SR $\uparrow$ \\
        \midrule
        UniVLA$^\dagger$      & $\mathbf{87.5{\pm}0.1}$ & $65.7{\pm}6.1$ & $35.2{\pm}4.7$ & $20.5{\pm}4.7$ & $70.3{\pm}5.6$ & $31.3{\pm}4.0$ & $43.6{\pm}2.4$ & $17.1{\pm}2.0$ \\
        MoDE$^\dagger$        & $69.7{\pm}1.8$ & $23.2{\pm}2.9$ & $46.8{\pm}2.9$ & $41.6{\pm}4.8$ & $71.1{\pm}1.5$ & $46.4{\pm}1.1$ & $32.9{\pm}1.4$ & $19.9{\pm}1.4$ \\
        T-Stellar$^\dagger$   & $76.6{\pm}3.4$ & $19.1{\pm}3.6$ & $57.5{\pm}1.1$ & $57.0{\pm}2.0$ & $80.3{\pm}1.6$ & $38.0{\pm}2.5$ & $48.6{\pm}2.6$ & $32.2{\pm}4.4$ \\
        TS-Stellar$^\dagger$  & $71.8{\pm}4.0$ & $25.9{\pm}2.7$ & $47.5{\pm}2.5$ & $47.7{\pm}1.4$ & $\mathbf{84.7{\pm}1.4}$ & $40.8{\pm}0.6$ & $51.4{\pm}1.4$ & $39.7{\pm}1.6$ \\
        IsCiL$^\dagger$ & $66.8{\pm}1.3$ & $7.8{\pm}2.5$ & $60.9{\pm}2.8$ & $48.5{\pm}2.0$ & $65.8{\pm}0.3$ & $31.5{\pm}0.3$ & $41.4{\pm}0.1$ & $17.3{\pm}0.3$ \\
        LOTUS$^\dagger$ & $46.8{\pm}4.3$ & $37.7{\pm}3.9$ & $18.5{\pm}1.5$ & $10.8{\pm}1.1$ & $32.3{\pm}2.3$ & $2.8{\pm}2.7$ & $29.4{\pm}0.5$ & $31.8{\pm}2.3$ \\
        \midrule
        Sequential   & $81.7{\pm}0.7$ & $74.1{\pm}0.6$ & $24.3{\pm}0.4$ & $7.5{\pm}0.1$ & $61.7{\pm}0.7$ & $58.1{\pm}0.3$ & $14.4{\pm}0.7$ & $3.7{\pm}0.6$ \\
        ER           & $74.7{\pm}0.4$ & $17.3{\pm}0.3$ & $60.5{\pm}0.3$ & $44.9{\pm}0.1$ & $73.3{\pm}1.4$ & $22.5{\pm}0.4$ & $55.2{\pm}1.2$ & $50.9{\pm}0.9$ \\
        Task-MoE     & $84.0{\pm}0.5$ & $14.4{\pm}1.2$ & $72.3{\pm}0.4$ & $66.0{\pm}1.1$ & $73.5{\pm}0.8$ & $12.1{\pm}0.5$ & $64.4{\pm}0.6$ & $61.1{\pm}2.9$ \\
        \midrule
        \rowcolor{gray!18}[\tabcolsep][\tabcolsep]
        SCE    & $86.5{\pm}0.3$ & $\mathbf{4.3{\pm}0.4}$ & $\mathbf{82.9{\pm}0.4}$ & $\mathbf{83.9{\pm}0.6}$ & $71.7{\pm}1.9$ & $\mathbf{2.5{\pm}1.8}$ & $\mathbf{70.2{\pm}0.2}$ & $\mathbf{73.9{\pm}1.0}$ \\
        \bottomrule
    \end{tabular}
    }
    
\end{table*}

\subsection{Simulation experiments}
\label{ssec:simulation}
We evaluate SCE on the LIBERO benchmark~\citep{liu2023libero}, including the goal-conditioned LIBERO-Goal and long-horizon LIBERO-Long. 

\textbf{Continual learning protocol.} Each suite forms a 10-stage stream in benchmark order, following Stellar VLA~\citep{wu2025stellarvla}. Unless specified otherwise, each stage trains on the new task and five replay demonstrations per old task. We evaluate all observed tasks after each stage, using 50 rollouts per task.

\textbf{Baselines.}
We compare with UniVLA, MoDE, T-Stellar, TS-Stellar, IsCiL, and LOTUS using results reported by Stellar VLA~\citep{wu2025stellarvla}. Our Sequential, ER, and Task-MoE implementations share the same pretrained $\pi_0$ backbone as SCE. Sequential and ER isolate replay; Task-MoE assigns one private expert per task.

\textbf{Metrics.}
We evaluate performance using four continual learning metrics: FWT, NBT, AUC, and Final SR.
Let $R_{t,k}$ denote the success rate after training through stage $t$ and evaluating the task introduced at stage $k$, where $1\leq k\leq t\leq T$ and $T$ is the number of continual stages.
FWT measures the diagonal performance when each task is first learned, computed as $\mathrm{FWT}=\frac{1}{T}\sum_{k=1}^{T}R_{k,k}$.
NBT measures later-stage degradation; following the LIBERO-style reporting convention used in our experiments, we compute $\mathrm{NBT}=\frac{1}{T}\sum_{k=1}^{T-1}\frac{1}{T-k}\sum_{t=k+1}^{T}(R_{k,k}-R_{t,k})$, where the final task has no forgetting term.
AUC averages each task's success-rate trajectory after its introduction, $\mathrm{AUC}=\frac{1}{T}\sum_{k=1}^{T}\frac{1}{T-k+1}\sum_{t=k}^{T}R_{t,k}$.
Final SR reports the average success rate over all tasks after the final stage, $\mathrm{Final\ SR}=\frac{1}{T}\sum_{k=1}^{T}R_{T,k}$.

\textbf{Results.}
Table~\ref{tab:main-results} shows that SCE achieves the best AUC, Final SR, and NBT on both primary suites. On LIBERO-Goal, it reaches 82.9 AUC and 83.9 Final SR, exceeding Task-MoE by 10.6 and 17.9 points. On LIBERO-Long, it improves AUC from 64.4 to 70.2 and Final SR from 61.1 to 73.9, while reducing NBT from 12.1 to 2.5. The matched-backbone comparison with Task-MoE highlights the value of organizing adaptation around reusable skills under the same replay protocol. Improvements in both AUC and Final SR show that the advantage extends across the learning stream and persists at the final stage, supporting skill composition for retaining previously acquired behaviors as new tasks are introduced.

\textbf{Replay efficiency.}
To evaluate replay efficiency, we compare SCE with ER using a larger replay buffer on LIBERO-Goal. Under the same update budget, SCE uses five replay demonstrations per old task and achieves 82.5 AUC and 83.4 Final SR, compared with 77.5 AUC and 79.0 Final SR for ER using 32 demonstrations per old task.

\textbf{Skill arrival order.}
We reorder the same ten LIBERO-Goal tasks, keeping the final five-skill set unchanged while varying when new skills enter the skill base. With new skills introduced at stages 6 and 9, SCE achieves 80.7 AUC and 81.4 Final SR, supporting continued learning as the skill base expands.

\newpage
\subsection{Real-world experiments}
\label{ssec:realworld}

\begin{figure}[t]
    \centering
    \includegraphics[width=\linewidth]{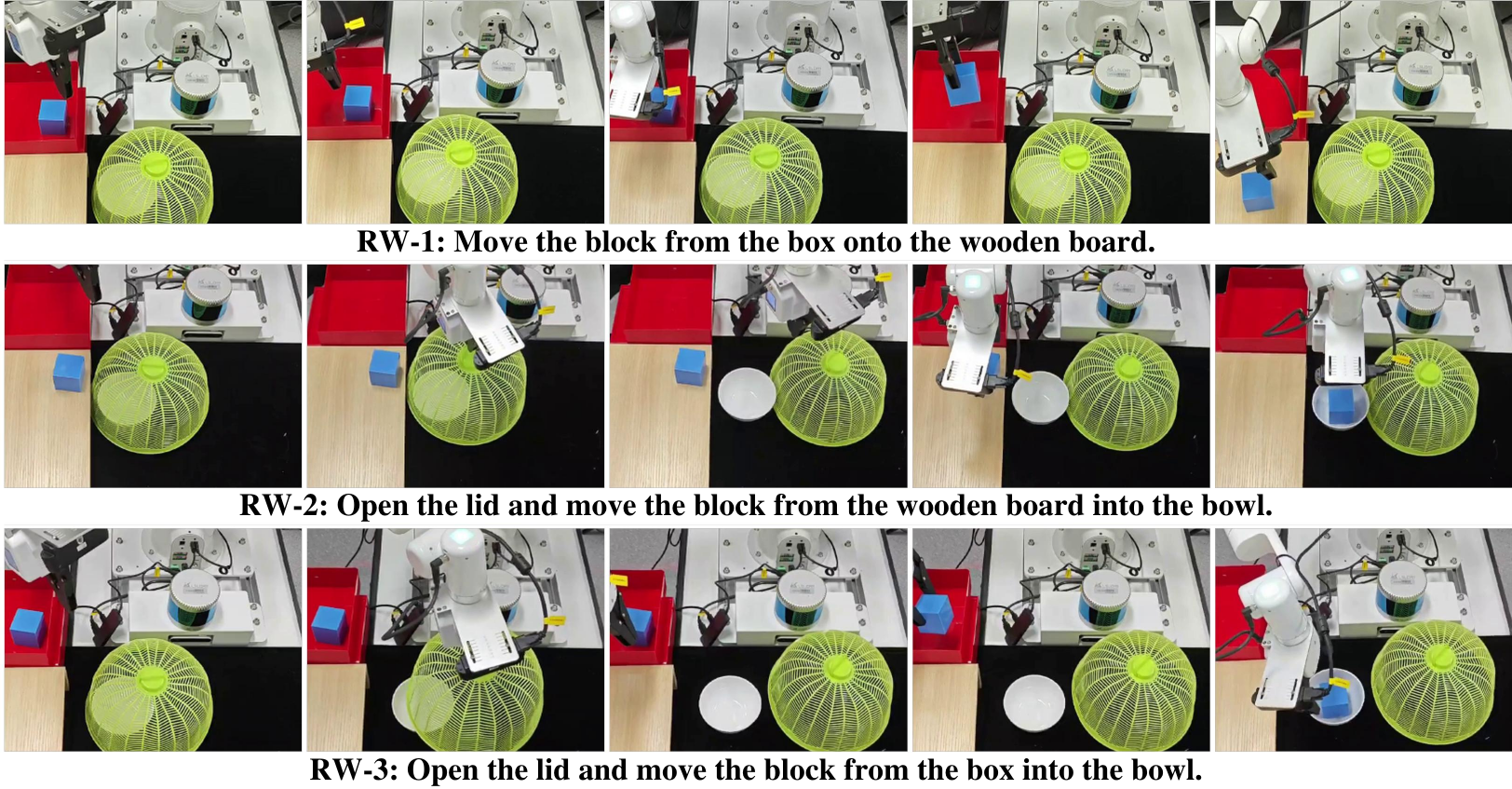}
    \caption{
      Representative snapshots of RW-1, RW-2, and RW-3, ordered from task start to completion.
      }
    \label{fig:realworld-overview}
\end{figure}

\begin{wraptable}{r}{0.52\textwidth}
    \centering
    \vspace{-6pt}
    \captionsetup{font=small}
    \caption{
    Real-world continual learning results.
    }
    \label{tab:realworld-results}
    {\small
    \setlength{\tabcolsep}{3pt}
    \begin{tabular}{lcccc}
        \toprule
        Method & FWT $\uparrow$ & NBT $\downarrow$ & AUC $\uparrow$ & Final SR $\uparrow$ \\
        \midrule
        Sequential & 55.6 & 45.6 & 28.7 & 10.0 \\
        ER         & 52.2 & 23.9 & 37.2 & 22.2 \\
        Task-MoE   & 64.4 & 8.3 & 59.8 & 58.9 \\
        \midrule
        \rowcolor{gray!18}[\tabcolsep][\tabcolsep]
        SCE       & \textbf{80.0} & \textbf{0.6} & \textbf{79.1} & \textbf{80.0} \\
        \bottomrule
    \end{tabular}
    }
\end{wraptable}
We evaluate the Mercury B1's 7-DoF arm on three sequential tasks: RW-1 transfers a block from a box to a board; RW-2 opens a lid and transfers the block from the board into a bowl; RW-3 opens the lid and transfers the block from the box into the bowl. These tasks test reuse of retrieval, transfer, placement, and lid-opening skills. Fig.~\ref{fig:realworld-overview} shows representative rollouts; videos on the \href{https://eqcy.github.io/sce/\#videos}{project website} show execution.

The three tasks share manipulation behaviors while varying the block's source and destination. In particular, RW-2 and RW-3 both require opening the lid and placing the block into the bowl, but retrieve it from different locations. This task stream evaluates reuse of learned behaviors across related manipulation tasks.

\textbf{Protocol and baselines.} SCE, Sequential, ER, and Task-MoE share the pretrained $\pi_0$ backbone and three-stage protocol. Replay retains five demonstrations per old task; each observed task is evaluated over 30 trials per method after each stage.

\textbf{Results.} As shown in Table~\ref{tab:realworld-results}, SCE leads all four metrics, improving Final SR from 58.9 to 80.0 and reducing NBT from 8.3 to 0.6 over Task-MoE. These results show that SCE's advantages also hold on real robots, enabling it to learn new tasks while retaining previously acquired behaviors.

\subsection{Analysis of feature drift impact under closed-loop control}
\label{ssec:drift_analysis}

\begin{figure}[t]
  \centering
  \includegraphics[width=0.86\linewidth]{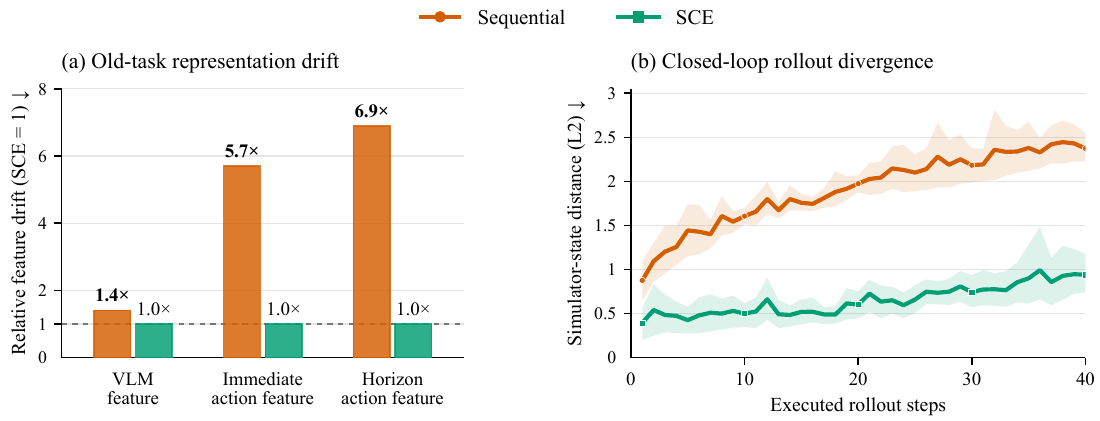}
  \caption{Old-task feature retention and rollout stability.
    (a) Mean feature cosine distance, normalized to SCE within each feature type.
    (b) Simulator-state L2 distance between each method's acquisition and final-checkpoint rollouts from shared initial states.
    Curves show task-averaged distances; shading denotes pointwise 95\% bootstrap confidence intervals across tasks. Lower is better in both panels.
    }
  \label{fig:drift-impact-closed-loop}
\end{figure}

SCE preserves old-task representations and rollout behavior. We compare acquisition and final checkpoints on identical demonstration inputs, measuring cosine distance in visual-language representations and decoder features before action projection. Immediate-action drift uses the first action position; horizon-action drift uses hidden features averaged across action positions. Fig.~\ref{fig:drift-impact-closed-loop}(a) shows that Sequential's drift is respectively 1.4, 5.7, and 6.9 times SCE's at these three levels. Fig.~\ref{fig:drift-impact-closed-loop}(b) further shows that SCE maintains smaller rollout state distances throughout execution when both checkpoints start from identical simulator states.

\newpage

\subsection{Ablation experiments}
\label{ssec:ablation}

\begin{wraptable}{r}{0.54\textwidth}
  \centering
  \vspace{-6pt}
  \captionsetup{font=small}
  \caption{Ablations on LIBERO-Goal. EEB/TEB: Execution/Transition Expert Branch.}
  \label{tab:ablation-libero-goal}
  {\small
  \setlength{\tabcolsep}{3pt}
  \begin{tabular}{@{}lcccc@{}}
      \toprule
      Method & FWT $\uparrow$ & NBT $\downarrow$ & AUC $\uparrow$ & Final SR $\uparrow$ \\
      \midrule
      SCE w/o CSG & 72.4 & 7.0 & 66.1 & 63.6 \\
      SCE w/o TEB & 61.4 & \textbf{-6.3} & 67.1 & 69.4 \\
      SCE w/o EEB & 80.6 & 6.5 & 75.0 & 80.0 \\
      \midrule
      \rowcolor{gray!18}
      SCE & \textbf{86.2} & 4.3 & \textbf{82.5} & \textbf{83.4} \\
      \bottomrule
  \end{tabular}
  }
\end{wraptable}
Table~\ref{tab:ablation-libero-goal} evaluates the contributions of CSG and DETE. Compared with SCE w/o CSG, full SCE improves AUC from 66.1 to 82.5 and Final SR from 63.6 to 83.4, demonstrating the benefit of explicitly organizing reusable skills.

Within DETE, adding the Transition Expert Branch improves FWT from 61.4 to 86.2 and AUC from 67.1 to 82.5, while adding the Execution Expert Branch improves AUC from 75.0 to 82.5 and reduces NBT from 6.5 to 4.3. These results support the complementary roles of skill-specific execution and cross-skill transitions in continual task learning.

Across LIBERO-Goal tasks, the Execution-branch fusion weight decreases by 16.2\% on average near skill boundaries, while the normalized contribution of the Transition Expert Branch increases in representative transition windows. These observations show how adaptive fusion shifts emphasis from skill-specific execution toward transition modeling during skill changes.

\section{Conclusion}

We propose SCE, a skill-compositional framework for ECL that constructs a reusable skill base from sequential task demonstrations through CSG and composes these skills for new task learning through DETE. By combining skill-specific execution with adaptive cross-skill transitions, SCE supports stable reuse of learned behaviors across evolving tasks. Experiments on LIBERO benchmarks and real-world manipulation tasks show improved retention and overall continual learning performance. Feature-drift analyses and ablation studies connect these gains to reduced representational drift and coherent skill composition under closed-loop control. A promising future direction is to extend skill-compositional ECL to larger and more open-ended skill repertoires.

\small
\setlength{\bibsep}{3pt}
\bibliographystyle{plain}
\bibliography{references}

\end{document}